\documentclass[letterpaper, 10pt, conference]{IEEEtran}
\ifCLASSINFOpdf
\else
\fi
\usepackage{amsmath}

\usepackage{graphicx}
\usepackage{subfigure}
\usepackage{subeqnarray}
\usepackage{cases,booktabs,tabu}
\usepackage{epsfig}
\usepackage{epstopdf}
\graphicspath{{figure/}{figures/}}
\begin{document}
%
\title{Motion Control on Bionic Eyes: A Comprehensive Review}

\author{%
\IEEEauthorblockN{Zheng Zhu, Qingbin Wang, Wei Zou, Feng Zhang}
}

\maketitle

\begin{abstract}
Biology can provide biomimetic components and new control principles for robotics. Developing a robot system equipped with bionic eyes is a difficult but exciting task. Researchers have been studying the control mechanisms of bionic eyes for many years and considerable models are available. In this paper, control model and its implementation on robots for bionic eyes are reviewed, which covers saccade, smooth pursuit, vergence, vestibule-ocular reflex (VOR), optokinetic reflex (OKR) and eye-head coordination. What is more, some problems and possible solutions in the field of bionic eyes are discussed and analyzed. This review paper can be used as a guide for researchers to identify potential research problems and solutions of the bionic eyes' motion control.
\end{abstract}

\IEEEpeerreviewmaketitle

\section{Introduction}
\label{sec1}
Eyes are the most important sensors for human beings during the process of information acquisition. More than 80 percent of information is acquired by eyes. The human visual system is highly developed and perfect after millions of years of evolution. Visual system can lock the object at the center of the retina (foveal area) even when the position of head or object changes drastically. This is of great significance for robots who always work in the bumpy and unstructured environment. Studying bionic eyes which can act like human beings is a difficult but exciting task. A robot system equipped with cameras has a better performance in terms of perception ~\cite{46} and control~\cite{47,48,53}, which is of significance for real-world applications such as visual tracking~\cite{44, 49, 51,52, 56}, moving object detection~\cite{57,a,b}, action recognition~\cite{45,50,54} and segmentation~\cite{55}.

Main motion forms of bionic eyes include saccade, smooth pursuit, vergence, vestibule-ocular reflex (V,,OR), optokinetic reflex (OKR) and eye-head coordination. Saccade is used to move eyes voluntarily from one point to another by rapid jumping, while smooth pursuit can be applied to track moving targets. VOR acts to stabilize retinal images by generating a compensatory eye motion during head turns. OKR can stabilize retinal images for gazing at rapid moving objects by nystagmus. OKR is driven by retinal slip while VOR is driven by head velocity signal. Commonly, two or more forms of motion work simultaneously. Besides, the binocular coordination and eye-head coordination are of high importance to realize object tracking and gaze control.

The control mechanisms of bionic eyes have been studied for many years. Besides, some robot systems equipped with bionic eyes have been designed to implement these control models. Some typical systems of them are listed as following: The iCub robot~\cite{1} has 3 DOFs on the neck and 3 DOFs for the eyes. The KOBIAN robot~\cite{2} has 3 DOFs on the eyes, and 4 DOFs on the neck. The ARMAR robot~\cite{3} is designed to study gaze control. The Romeo~\cite{4} is a humanoid robot which employs 4 DOFs in the eyes. There are also some robot head systems such as the BARTHOC head~\cite{5} and the Flobi head~\cite{6} et al. Implementations of bionic eyes' motion control mechanisms on these robot systems have validated their effectiveness for improving perception performance. However, restricted to the development of neuroscience, there are difficulties when imitating the performance of human eyes.

The purpose of studying bionics is to imitate eye movements of primates to get a better performance in many aspects, such as gaze shifts and image stabilization. Most previous overviews about bionic eyes focused on the study of neurophysiology and designed control models to imitate these behaviors, but implementations of these control models on robot systems are always ignored. So, besides the current state-of-the-art of bionic eyes' control mechanisms, their implementations on robot system are also discussed and analyzed in this paper. This review paper can be used as a guide for researchers to identify potential benefits and limitations of the bionic eyes' study.

The paper is organized as follows: In Section~\ref{sec2}, research status of different forms of bionic eyes' motion is summarized, including saccade, smooth pursuit, VOR, OKR and eye-head coordination. Various models and their features are highlighted. What is more, implementations of control models on robots are presented. In Section~\ref{sec3}, some problems and possible solutions in the field of bionic eyes are discussed and analyzed. Finally, conclusions are drawn in Section~IV.

\section{Overview of Motion Control on Bionic Eyes}
\label{sec2}
Biology can provide biomimetic components and new control principles for robotics. The motion forms of primate eyes include saccade, smooth pursuit, vergence, VOR, OKR and eye-head coordination. Studying on these motion forms can help researchers to improve the performance of the robots' visual systems, including image stabilization, object tracking, navigation, and so on. Thanks to the efforts of many researchers, considerable control models for bionic eyes and their implementations are available.

\subsection{Models of Saccade}
Saccade is used to move eyes voluntarily from one point to another by rapid jumping. It is of great significance for robots to change their fixation point quickly. In the control models, saccade control system should act as a position servo controller to change and keep the target at the center of the retina with minimum time consuming.

\begin{figure}[!ht]
\centerline{\includegraphics[width=\columnwidth]{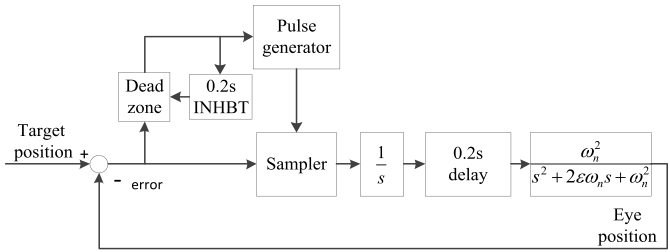}}
\caption{Young's model for saccade, see text for symbols.}
\label{fig1}
\end{figure}

\begin{figure}[!ht]
\centerline{\includegraphics[width=\columnwidth]{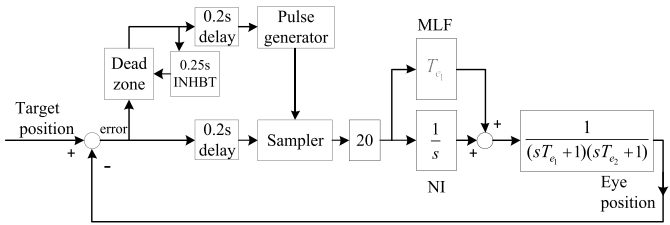}}
\caption{Robinson's model for saccade, see text for symbols.}
\label{fig2}
\end{figure}

Young and Stark~\cite{7} proposed the sampled data model for saccade in 1963, which is shown in Fig.~\ref{fig1}. The circuit contains the dead zone and the INHBT (a device to inhibit the timing circuit), when the error exceeds a certain threshold, the pulse generator is triggered which causes a sample to be taken, at the same time, the INHBT element blocks dead zone for 0.2 seconds. The proposed sampled data model describes saccade by using discrete rather than continuous control loops. This model didn't take into account the actual brain structures, because so little was known at that time about the brain stem organization of eye movements. To solve this problem, Robinson~\cite{8} modified Young's model by adding the premotor circuitry, which consists of medial longitudinal fasciculus (MLF) and neural integrator (NI). As shown in Fig.~\ref{fig2}, the MLF forms a lead network to compensate the lag of the plant, while the NI produces the position signal. Then the sum of MLF and NI pathway can be regarded as the input of the eyeball mechanics. The modifications are consistent with neurophysiology because of the fact that brain processes data in parallel.

\begin{figure}[!ht]
\centerline{\includegraphics[width=.6\columnwidth]{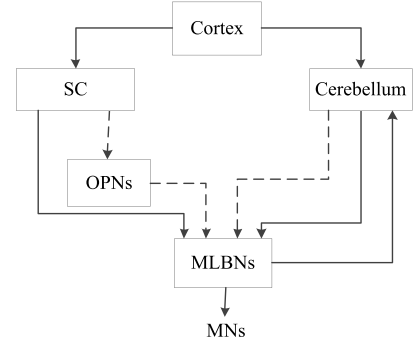}}
\caption{Quaia's model for saccade, see text for symbols.}
\label{fig3}
\end{figure}

Both clinical and experimental evidences indicate that the superior colliculus (SC) and the cerebellum are important to produce accurate saccadic movements, and Quaia et al.~\cite{9} proposed a new model for saccade based on this fact. As shown in Fig.~\ref{fig3}, the solid line represents excitatory signals, the dotted line represents inhibitory signals, OPNs represents omnipause neurons, and MLBNs represents medium lead burst neurons. The superior colliculus pathway is parallel with the cerebellum pathway. The superior colliculus pathway provides a directional drive of eye movement while the cerebellum pathway keeps track of the saccade toward the target. Sum of these two drives is passed on to the motoneurons (MNs) and determines the velocity of the eyes.

Besides, researchers have been studying the implementations of saccade on robots in the last twenty years. In 1997, Bruske et al.~\cite{10} implemented saccadic control on a binocular vision system by using the feedback error learning (FEL) strategy. In 2013, Wang et al.~\cite{11} designed an active vision system which can imitate saccade and other eye movements. The saccadic movements are implemented with open-loop controller, which ensures faster saccade eye movements compared with closed-loop controller. In 2015, Antonelli et al.~\cite{12} realized saccadic movements on a robot head by using a model called recurrent architecture (RA). In this model, the cerebellum is regarded as an adaptive element to learn internal model while brainstem is regarded as a fixed-inverse model. The experimental results on robot show that this model is more accurate and less sensitive to the choice of the inverse model compared with the FEL model. These models or methods for saccade and their features are summarized in Table~\ref{tab1}.
\begin{table}[!ht]
\centering
\caption{Models or methods for saccade and their features see text for symbols}
\tabcolsep=3pt
\tabulinesep=2pt
\begin{tabu}to\columnwidth{X[1cm]X[1cm]X[1cm]}
\toprule
Time and authors&
Models or methods&
Features \\
\midrule
1963, Young and Stark&
Sampled data model&
Discrete control loops, no brain structures \\
1973, Robinson&
MLF and NI model&
Based on the Young's model \\
1997, Bruske et al.&
The FEL model&
Fast and accurate, implemented on the robot \\
1999, Quaia et al.&
SC and cerebellum model&
The SC and cerebellum produce the saccade \\
2013, Wang et al.&
Open-loop controller&
Fast, implemented on the robot \\
2015, Antonelli et al.&
Recurrent architecture&
More accurate and less sensitive, implemented on the robot \\
\bottomrule
\end{tabu}
\label{tab1}
\end{table}

Study on saccadic movements has the longest history in control of bionic eyes. Saccade should act as a position servo controller to change and keep the target at the center of the retina. This is of great significance for bionic eyes to shift gaze quickly. Implementations of saccade on robots have been studied in the last twenty years. According to the implemented approaches, control models are designed to imitate the functions of brain regions such as superior colliculus, cerebellum and brainstem.

\subsection{Models of Smooth Pursuit}

\begin{figure}[!ht]
\centerline{\includegraphics[width=\columnwidth]{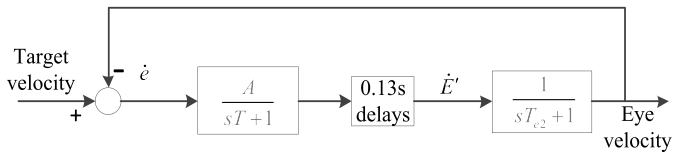}}
\caption{A simple model of pursuit, see text for symbols.}
\label{fig4}
\end{figure}

Smooth pursuit is performed to track an object moving smoothly. Robots can track moving targets better by imitating the characteristics of smooth pursuit. In the control models, smooth pursuit system should act as a velocity servo controller to rotate the eyes at the same angular rate as the target. Fig.~\ref{fig4} illustrates a simple model of pursuit~\cite{13}, the velocity of the target's image across the retina, $\dot{e}$, is taken as the major stimulus to pursuit and the error that the system tries to minimize; then is transformed into an eye velocity command, $\dot{E}'$, to be sent to motoneurons.

In 1986, the study of Robinson et al.~\cite{13} showed that smooth pursuit is a continuous negative feedback system responding to the eyeball velocity, which is different from saccade. Many current smooth pursuit models are descendants of the original model of Robinson. Based on the Robinson's model, Brown~\cite{14} added Smith predictor to cope with delays in loop control. In 1987, Lisberger et al.~\cite{15} proposed a model for smooth pursuit. The input of this model is the motion signal of retina. Then the motion signal is used to produce the position signal, velocity signal and acceleration signal of retina. Those signals are weighted and sent to an integrator to produce the velocity command of eye movements. In 1989, Deno et al.~\cite{16} applied dynamical neural network, which helps to unify two apparently disparate models of smooth pursuit and clarify dynamical element organization, to the smooth pursuit system. The dynamical neural network can compensate delays from sensory input to motor response. In 1998, Lunghi et al.~\cite{17} introduced a neural adaptive predictor which is previously trained to accomplish smooth pursuit. This model can explain human's ability to compensate the 130ms physiological delays when human beings follow the external targets with their eyes.

Besides, some smooth pursuit models have been validated on the systems of bionic eyes. In 2001, Shibata et al.~\cite{18} implemented smooth pursuit control on a humanoid head. The control model can be divided into three subsystems: an inverse model controller, a target velocity predictor and a feedback controller. The results on robot head show a rapid convergence when learning the target dynamics. In 2012, Song and Zhang~\cite{19} proposed a binocular control model, which is derived from neural pathway, for smooth pursuit and other eye movements. In their smooth pursuit experiments, the maximum retinal error is less than 2.2 degrees, which can keep the target in the field of view accurately. In 2013, Wang et al.~\cite{11} implemented smooth pursuit with velocity closed-loop controller, which is based on the image information. Their experiments showed that the controller can adjust the eye velocity according to the distance between the target and the image center. These models or methods for the smooth pursuit and their features are summarized in Table~\ref{tab2}.
\begin{table}[!ht]
\centering
\caption{Models or methods for the smooth pursuit and their features}
\tabcolsep=3pt
\tabulinesep=2pt
\begin{tabu}to\columnwidth{X[1cm]X[1cm]X[1cm]}
\toprule
Time and authors&
Models or methods&
Features \\
\midrule
1986, Robinson et al.&
The velocity negative feedback model&
Original model for smooth pursuit \\
1987, Lisberger et al.&
Weighted signals&
Solves the conflict between high gain and large delay \\
1989, Deno et al.&
The dynamical neural network model&
Adaptive \\
1990, Brown&
Smith~predictor&
Based on Robinson's model \\
1998, Lunghi et al.&
Neural adaptive predictor&
Explain the compensation of delays \\
2001 Shibata et al.&
Be divided three subsystems&
Rapid convergence \\
2012, Song and Zhang&
Derived from neural pathways&
Integrated, implemented on the robot \\
2013, Wang et al.&
Closed-loop controller&
Adjust the eye velocity \\
\bottomrule
\end{tabu}
\label{tab2}
\end{table}

Realized approaches show that control models for smooth pursuit should use some predictive controllers. From the cybernetic point of view, how to solve the delay problem is of great significance. Besides, coordination of smooth pursuit and other eye movements are worthy of further investigation.

\subsection{Models of VOR and OKR}
VOR acts to stabilize the retinal images by generating a compensatory eye motion during head turns. OKR occurs when the target moves rapidly in front of the eyes and it is driven by the retinal slip. OKR comes in two stages, a faster (saccade) and a slower (smooth pursuit).

\begin{figure}[!ht]
\centerline{\includegraphics[width=\columnwidth]{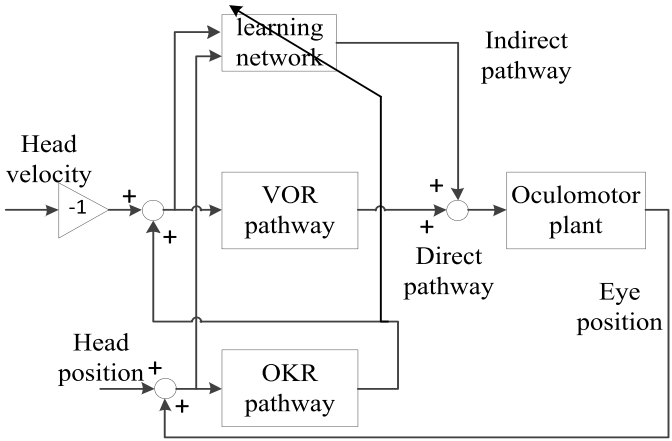}}
\caption{The FEL model with learning network.}
\label{fig5}
\end{figure}

In 1992, Gomi and Kawato~\cite{20} presented an adaptive feedback control and feedforward control model for VOR-OKR. This model is based on the feedback error learning (FEL) scheme. By this method, it is concluded that the learning is accomplished in cerebellum region. In 1998, Raymond and Lisberger~\cite{21} found that the brainstem~also plays an important role in the learning process. In 2001, Shibata and Schaal~\cite{22} added a learning controller as an indirect pathway for the VOR-OKR system. As shown in Fig.~\ref{fig5}, the learning controller takes the desired velocity and the estimated desired position as the input, and outputs necessary feedforward torque. VOR system is defined as a feedforward open-loop controller with an inverse model. OKR system is defined as a compensatory negative feedback controller for the VOR. This model can acquire good VOR performance after about 10 seconds of learning. Furthermore, it could converge to excellent performance after about 30 to 40 seconds of learning.

According to the clinical evidence, VOR includes rotational VOR (rVOR) and translational VOR (tVOR). They are controlled by semicircular canals and otolith organs, respectively. In 2005, Merfeld et al.~\cite{23} and Ramat et al.~\cite{24} proposed internal model and simple filtering model for rVOR and tVOR by clinical experiments, respectively. Porrill et al.~\cite{25} proposed a decorrelation control model for the VOR-OKR in 2004. The model is to remove any correlation between motor command and the variable that codes sensory error. It is worth noting that this model doesn't need the motor error signal by proposing a recurrent architecture. In 2010, Franchi et al.~\cite{1} compared the FEL and decorrelation model, and concluded that main difference between them is the learning control strategy. In the FEL model, a learning network is important and VOR collaborates with OKR sharing the frequency bandwidth response. In the decorrelation model, the OKR component is not taken into account and the model mainly focuses on the learning of cerebellar.

In 2007, Khojasteh and Galiana~\cite{26} found that there are human's vergence eye movements during the VOR in the dark. It was believed that the eyes move in a perfectly conjugate fashion during the VOR. But Khojasteh et al. observed a significant vergence movement that is modulated with head velocity during VOR. This observation suggests a vestibular contribution to vergence.

Many VOR-OKR control models have been validated on robot systems equipped with bionic eyes. In 2010, Kwon et al.~\cite{27} implemented the VOR-OKR-based vision tracking system on the mobile robot. The VOR-OKR concept is realized by using the robot's motion information from the artificial vestibular system (AVS) sensor cluster and the vision information from the cameras. In 2012, Song and Zhang~\cite{19} implemented VOR-OKR movements on their binocular robot head by deriving the control model from neural pathways and pre-existing binocular models. Experiments show that better image stabilization result can be obtained by combining the VOR-OKR movements and the smooth pursuit movements. In 2013, Wang et al.~\cite{11} imitated the VOR movements using PID controller. And the angular velocity is obtained by placing an inertia tracker at the rotational axis of the head. Experimental results show that VOR can shorten the time of reaching to the target compared with eyes movements without VOR, which is from 5 seconds to 2.75 seconds. These models or methods for the VOR-OKR and their features are summarized in Table~\ref{tab3}.
\begin{table}[!ht]
\centering
\caption{Models or methods for the VOR-OKR and their features}
\tabcolsep=3pt
\tabulinesep=2pt
\begin{tabu}to\columnwidth{X[1cm]X[1cm]X[1cm]}
\toprule
Time and authors&
Models or methods&
Features \\
\midrule
1992, Gomi and Kawato&
Feedback and feedforward control&
Based on FEL \\
2001, Shibata and Schaal&
FEL model with learning network.&
Add a learning controller as an indirect pathway \\
2004, Porrill et al.&
Decorrelation control model&
Minimize the correlation function \\
2005, Merfeld et al. and Ramat et al.&
Internal model and simple filtering model&
Explain the rVOR and tVOR \\
2010 Kwon et al.&
Use the robot motion information and vision information&
Stable, robust, implemented on the robot \\
2012 Song and Zhang&
Combined with the smooth pursuit&
Can obtain better image stabilization result \\
2013 Wang et al.&
Use the inertia trackers&
Shorten the time of reaching to the target \\
\bottomrule
\end{tabu}
\label{tab3}
\end{table}

Image stabilization is a necessary ability for bionic eyes. Many models and methods of VOR-OKR are proposed to realize image stabilization. According to the above researches, the learning process plays an important role in VOR-OKR control models. Besides, better image stabilization can be obtained by combining VOR-OKR and other eye movements.

\subsection{Eye-head Coordination during the Gaze Shifts}

Human beings shift fixation by changing their head's or body's pose when the gaze shifts' amplitude is too large, because it is not enough to keep target at the center of the retina only by saccade. There are two problems about the eye-head coordination: What areas in the brain play an important role for the eye-head coordination? What new models can be proposed to explain the gaze shifts better?

For the first question, Freedman and Sparks~\cite{28} found that the desired gaze displacement command can be decomposed into separate eye and head displacement signals by the superior colliculus. Wang et al.~\cite{29} proposed a model that the gaze displacement signal is decomposed into eye and head contribution within the cerebellum. According to the above researches, both superior colliculus and cerebellum contribute to the eye-head coordination.

For the second question, the optimal control is an efficient method to explain eye-head coordination during the gaze shifts. Gaze shifts should be as fast as possible in order to increase the time that the image is stabilized on the retina. So, Enderle and Wolfe~\cite{30} proposed an optimal control method that minimizes the time to reach the target, i.e. the minimum time rule. Harris and Wolpert~\cite{31} suggested another optimality principle called minimum variance rule, which minimizes the variance of the eye position. Kardamakis and Moschovakis~\cite{32} proposed a new method to understand the eye-head gaze shifts with the help of optimal control theory. They found that the gaze shifts obey a simple physical principle, i.e. the minimum effort rule. In this model, the squared sum of the eye and the head torque signals integrated over the movement period are minimized in order to obtain the optimal control signal. In 2011, Saeb et al.~\cite{33} proposed another optimal control method for eye-head coordination. The method combines the incremental learning with the optimality principle. This model can realize local adaption mechanism by minimizing the defined cost function.

Many researchers have validated the eye-head coordination models on the systems of bionic eyes. In 2006, Maini et al.~\cite{34} implemented bionic eye-head coordination on their robot head. The control model used in the experiment is based on the methods in~\cite{35}. Both the eye and the head motors share the gaze displacement feedback signal while the gaze control system is driven by the collicular displacement signal. In 2011, Kido et al.~\cite{36} built a 7 DOFs robot to study eye-head coordination by gazing the moving targets. The control model is based on the methods in~\cite{37}. This model combines the eye-prioritized method and a mechanism that the head tends to return to the central position. In 2014, Vannucci et al.~\cite{38} proposed an adaptive controller to realize eye-head coordination with prediction of a moving target. The control model is improved based on the methods in~\cite{39}, which combine the growing neural gas and the motor babbling technique. These models or methods for the eye-head coordination and their features are summarized in Table~\ref{tab4}.
\begin{table}[!ht]
\centering
\caption{Models or methods for the eye-head coordination and their features}
\tabcolsep=2pt
\tabulinesep=2pt
\begin{tabu}to\columnwidth{X[1cm]X[1cm]X[1cm]}
\toprule
Time and authors&
Models or methods&
Features \\
\midrule
1987 Enderle and Wolfe&
Minimum time rule&
Minimize the time to reach the target \\
1998 Harris and Wolpert&
Minimum variance rule&
Realize local adaption \\
2006 Maini et al.&
Gaze control model based on paper~\cite{35}&
Low residual errors, no residual oscillations of the head \\
2009 Kardamakis and Moschovakis&
Minimum effort rule&
Combine optimal control and system modeling of neural processes \\
2011, Saeb et al.&
Incremental learning with optimality&
Minimize the cost function \\
2011 Kido et al.&
Control model based on paper~\cite{37}&
Simple, easy to implement \\
2014, Vannucci et al.&
Improved based on paper~\cite{39}&
Fast, accurate, implemented on the robots \\
\bottomrule
\end{tabu}
\label{tab4}
\end{table}

According to the above researches, both superior colliculus and cerebellum contribute to the eye-head coordination. The optimal control is an efficient method to explain eye-head coordination during the gaze shifts. Researchers have proposed some optimal control models based on different principles. Besides, how to implement these optimal control models should be paid more attention.

\section{Problems and Possible Solutions}
\label{sec3}
Researchers have been studying the control mechanisms of bionic eyes for many years. Thanks to the efforts of these researchers, considerable control models for bionic eyes and their implementations are available. However, there are still many difficulties and problems because of two reasons: firstly, the study of bionic eyes is restricted by the slow development of physiology and neuroscience. Because most current control models for bionic eyes are derived from neural pathways, but the neural mechanisms are too sophisticated to imitate perfectly. Secondly, there are common challenges in the robot field such as speed, accuracy and robustness. In this section, some problems and possible solutions for bionic eyes are discussed.

\subsection{Information Fusion of the Monocular Vision and Binocular Vision}
Depth and motion information can help developing qualified behavior for bionic eyes. The platforms of bionic eyes are always equipped with two cameras, so perception methods based on binocular vision and those based on monocular vision can be applied to reconstruct depth and motion information for bionic eyes, simultaneously. The perception accuracy of binocular vision relies on the extrinsic parameters between two cameras, while the monocular vision method relies on motion parameters. Information fusion of the monocular vision and binocular vision may be an efficient way to improve the accuracy of perception. To the best of our knowledge, this has not been attempted before.

\subsection{Problems of the Eye-head Coordination}

It is of high importance for bionic eyes to change fixation and keep the target on the fovea. The eye-head coordination is necessary to realize this goal. Many current models for eye-head coordination can't be applied to real robot systems directly because the compensatory mechanism of eyeballs is not clear completely. In 2011, Milighetti et al.~\cite{40} derived kinematic mathematical models for eye-head coordination, which is a general control algorithm without neural model. But the calculation of inverse kinematics equation is too complicated to be accomplished in acceptable time. How to realize eye-head coordination rapidly and accurately is still worthy of noting. Besides, combination of saccade and VOR during the eye-head coordination is also worthy of further investigation.

\subsection{Problems of the Binocular Coordination}

Almost all of primates have binocular systems and many robot systems are also equipped with binocular vision to tracking targets. But there are only a little researchers take into account the binocular coordination. Most control models regard binocular vision systems as two unrelated cameras and control them independently. The binocular coordination can bring additional information for the control of bionic eyes. For example, one of the cameras loses target at a certain time while the other camera still captures target at the same time, then the first camera can be redirected to the target easier with the information provided by the other camera. How to realize binocular coordination and use these additional information are worthy of consideration.

\subsection{Microsaccade}

When humans try to fix their gaze on a certain target, the eyeballs are not stationary completely. There are also eyes movements called microsaccades, which typically occur during prolonged visual fixation. They are small-amplitude and involuntary eye movements just like miniature versions of saccades. In the last decade, researchers have been studying microsaccades. Evidences indicate that microsaccades contain depth information and can highlight edge information. This may provide a new approach to bionic eyes to get a better performance in image processing. Until now, microsaccade is still a largely unsolved topic and this movement is worthy of further investigation.

\subsection{New Methods}
Advances in many interdisciplinary fields, including optimal control, intelligence control and machine learning, are of great significance for the research of bionic eyes. The optimal control theory is an essential way to solve eye-head coordination problems, which can always obtain a better performance compared with the traditional PID control methods. Neural networks and other machine learning methods can also be applied to train models of bionic eyes. It is worth noting that some researchers have been studying bionic compound eyes which imitate insect eyes~\cite{41,42,43}. Insect compound eyes have wider field of view and faster speed compared with human beings. This study may provide new approaches for the design of bionic eyes.

\section*{Conclusions}
\label{sec4}
Control mechanisms and implementations for bionic eyes have been studied for many years and considerable models are available. In this paper, control models and their implementations for bionic eyes are reviewed. The research status of motion forms is summarized. Implementations of control models are presented as well. What is more, this paper discusses some problems and possible solutions in the field of bionic eyes. This review paper can be used as a guide for researchers to identify potential research problems and solutions of the bionic eyes' motion control.

\bibliographystyle{IEEEtran}
\bibliography{mycite}

\end{document}